\documentclass{article}

\usepackage[utf8]{inputenc} 
\usepackage[T1]{fontenc}    
\usepackage{hyperref}       
\usepackage{url}            
\usepackage{booktabs}       
\usepackage{amsfonts}       
\usepackage{nicefrac}       
\usepackage{microtype}      
\usepackage{lipsum}
\usepackage{xcolor}
\usepackage{orcidlink}
\usepackage{fancyhdr}       
\usepackage{graphicx}       
\graphicspath{{media/}}     
\usepackage{wrapfig}
\usepackage{algorithm}
\usepackage{algpseudocode}
\definecolor{shadecolor}{RGB}{230,230,230}
\title{Duplicate Detection with GenAI}

\author{
  Ian Ormesher \\
  ZEISS Digital Partners \\
  Munich \\
  \texttt{ian.ormesher@zeiss.com} \\
}

\begin{document}
\maketitle

\begin{abstract}

Customer data is often stored as records in Customer Relations Management systems (CRMs). Data which is manually entered into such systems by one of more users over time leads to data replication, partial duplication or fuzzy duplication. This in turn means that there no longer a single source of truth for customers, contacts, accounts, etc. Downstream business processes become increasing complex and contrived without a unique mapping between a record in a CRM and the target customer. Current methods to detect and de-duplicate records use traditional Natural Language Processing techniques known as “Entity Matching”. In this paper we show how using the latest advancements in Large Language Models and Generative AI can vastly improve the identification and repair of duplicated records. On common benchmark datasets we find an improvement in the accuracy of data de-duplication  rates from 30\% using NLP techniques to almost 60\% using our proposed method.
\end{abstract}

\section{Introduction}
The task of identifying duplicate records is often done by pairwise record comparisons and is referred to as “Entity Matching” (EM). Whilst EM is broader in it’s scope and definition, we describe a technique here that can be used to identify duplicate records within a table using embedding vectors and distance metrics.

\section{Traditional Approach}
\label{sec:headings}

The five steps of a typical duplicate detection pipeline based on pairwise record comparisons (EM) is described by Panse et al. \cite{dupdeteval}.

This detection pipeline consists of several steps which include the following:
\begin{itemize}
\item Candidate Generation
\item Blocking
\item Matching
\item Clustering
\end{itemize}

\begin{figure}[h]
\includegraphics[width=\textwidth]{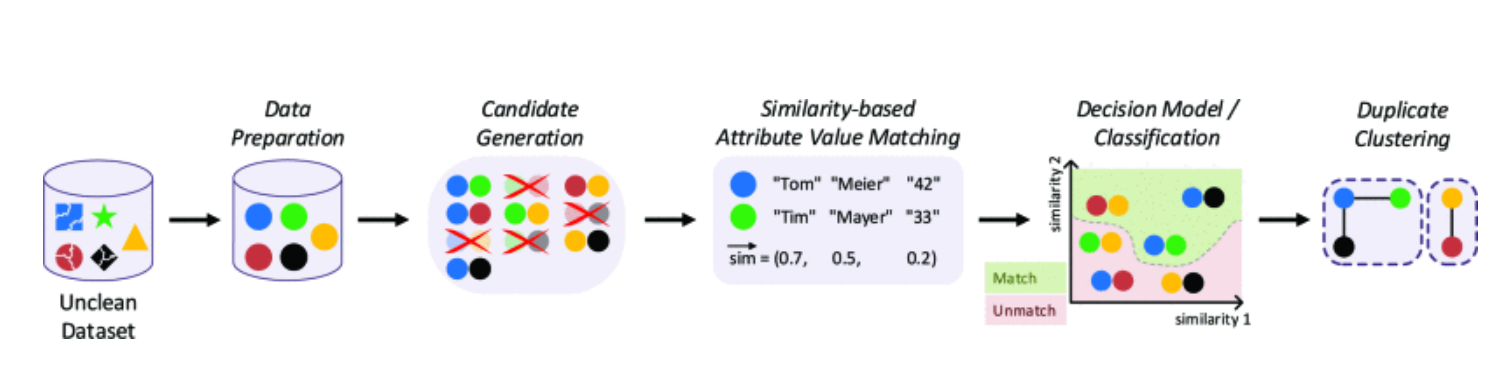}
\caption{The five steps of a typical duplicate detection pipeline based on pairwise record comparisons}
\label{fig:trad-em-label}
\end{figure}

These steps are illustrated by their diagram shown in Figure \ref{fig:trad-em-label}.

\subsection{Candidate Generation}
In the usual EM method, we would produce candidate records by combining all the records in the table with themselves to produce a cartesian product. For example, \(row _{i}\) is combined with \(row _{j}\) and produces a candidate record of \((row _{i} + row _{j})\). This process is repeated for all the rows with all the other rows. For a table of N rows this would produce (N x N) – N candidate records. We subtract N from the total because we never combine a row with itself. As an example, a table of 200 records would produce (200 x 200) – 200 = 39,800 candidate records.

\subsection{Blocking}
The idea of blocking is to eliminate those records that we know could not be duplicates of each other because they have different values for the “blocked” column. As an example, If we were considering customer records, a potential column to block on could be something like “City”. This is because we know that even if all the other details of the record are similar enough, they cannot be the same customer if they’re located in different cities. Once we have generated our candidate records, we then use blocking to eliminate those records that have different values for the blocked column.

\subsection{Matching}
Following on from blocking we now examine all the candidate records and calculate traditional NLP similarity-based attribute value metrics with the fields from \(row _{i}\) and \(row _{j}\). Using these metrics, we can determine if we have a potential match or un-match.

\subsection{Clustering}
Now that we have a list of candidate records that match, we can then group them into clusters.

\section{Proposed Method}
\label{sec:proposedmethod}
There are several steps to the proposed method, but the most important thing to note is that we no longer need to perform the “Candidate Generation” step of the traditional method. The new steps become:
\begin{itemize}
\item Create Match Sentences
\item Create Embedding Vectors
\item Clustering
\end{itemize}

\subsection{Create Match Sentences}
First a “Match Sentence” is created by concatenating the attributes we are interested in and separating them with spaces.

As an example, let's say that we have a customer record which looks like the one shown in Table \ref{tab:examplecustomerrecord}. 

\begin{table}[h]
 \caption{Example Customer Record}
  \begin{tabular}{|l|l|l|l|l|}
    \hline
    \textbf{name1} & \textbf{name2} & \textbf{name3} & \textbf{address} & \textbf{city} \\
    \hline\hline
    John & Hartley & Smith & 20 Main Street & London \\
    \hline
  \end{tabular}
  \label{tab:examplecustomerrecord}
\end{table}

We would create a "Match Sentence" by concatenating with spaces the \textbf{name1}, \textbf{name2}, \textbf{name3}, \textbf{address} and \textbf{city} attributes to make the following:

\colorbox{shadecolor}{John Hartley Smith 20 Main Street London}

\subsection{Create Embedding Vectors}
\begin{figure}[h]
\includegraphics[width=\textwidth]{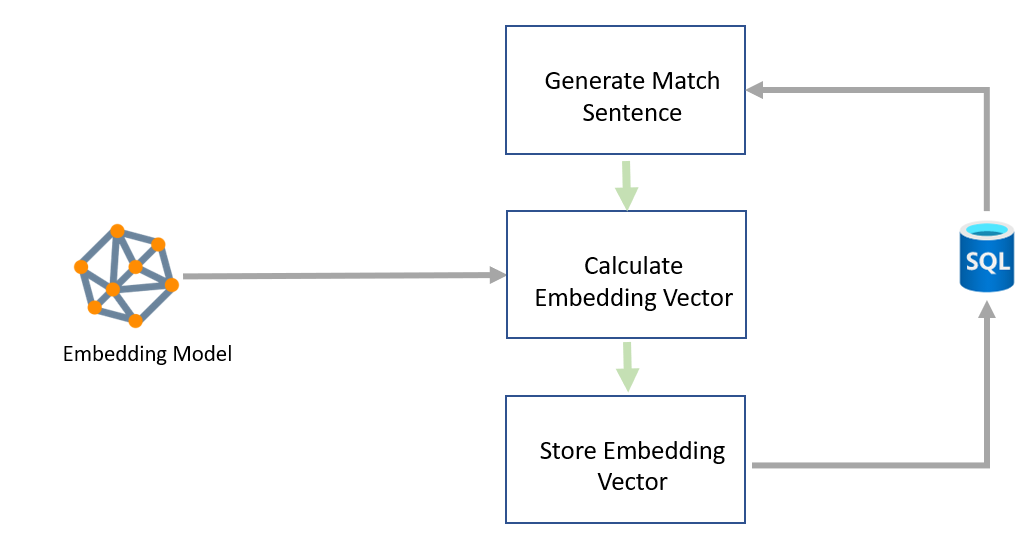}
\caption{The steps to creating the embedding vector}
\label{fig:create-embedding-vector}
\end{figure}

Once our “Match Sentence” has been created it is then encoded into vector space using our chosen embedding model. This is achieved by using “Sentence Transformers” \cite{sentence-transformers}. The output of this encoding will be a floating-point vector of pre-defined dimensions. These dimensions relate to the embedding model that is used. We used the \textbf{all-mpnet-base-v2}\cite{allmpnetbasev2} embedding model which has a vector space of 768 dimensions. This embedding vector is then appended to the record. This is done for all the records, as shown in Figure \ref{fig:create-embedding-vector}.

\subsection{Clustering}
Once embedding vectors have been calculated for all the records, the next step is to create clusters of similar records. To do this we use the DBSCAN technique.
DBSCAN works by first selecting a random record and finding records that are close to it using a distance metric. There are 2 different kinds of distance metrics that were used:

\begin{itemize}
\item L2 Norm distance
\item Cosine Similarity
\end{itemize}

For each of these metrics an epsilon value was used as a threshold value. All records that were within the epsilon distance and had the same value for the “blocked” column were then added to this cluster.
Once that cluster was complete another random record was selected from the unvisited records and a cluster then created around it. This then continues until all the records have been visited.
See the description of this algorithm using pseudo code \ref{alg:dbscan}.

\begin{algorithm}
\caption{DBSCAN}\label{alg:dbscan}
\begin{algorithmic}[1]
\Require $epsilon \geq 0$, $distancemetric$, $blockedcolumn(s)$
\State Pick a random unvisited record
\State Set this record state to visited and look for another unvisited record with the same value for the $blockedcolumn(s)$ and within $epsilon$ distance using the $distancemetric$
\If{no point is found}
    \State go back to step 1
\Else
    \State designate this a cluster group and add the record to it
    \State go back to step 2
\EndIf
\end{algorithmic}
\end{algorithm}

\section{Experiments}
\label{sec:experiments}
There were two sets of experiments that were performed:
\begin{itemize}
\item with our existing customer data
\item with the musicbrainz data
\end{itemize}

\subsection{Customer Data Experiments}
For each of the 2 distance measures various epsilon values were tried to see how this affected the clustering. They were also compared against the base traditional Entity Matching. See Table \ref{tab:experiments}.

\begin{table}[h]
 \caption{Customer Data Experiments}
  \centering
  \begin{tabular}{lll}
    \toprule
    ID & Distance Metric             & Epsilon \\
    \midrule
    1  & Traditional Entity Matching &         \\
    2  & Cosine similarity           & 0.1     \\
    3  & Cosine similarity           & 0.05    \\
    4  & L2 norm                     & 0.5     \\
    5  & L2 norm                     & 0.25    \\
    6  & L2 norm                     & 0.125   \\
    7  & Cosine similarity           & 0.025   \\
    \bottomrule
  \end{tabular}
  \label{tab:experiments}
\end{table}

The embedding model used was the \textbf{all-mpnet-base-v2}\cite{allmpnetbasev2} which encodes text into an embedding vector of 768 dimensions.

\subsubsection{Results}
\label{sec:results}
The values of epsilon for each of the different experiments were chosen based on the number of matching groups it produced and the size of the largest group. Since there is initially no ground truth with which to compare these records to (until the users label the data) we cannot show precision, accuracy or F-scores. Bearing that in mind, we can show the max group size and the nunber of match groups. These were good indicators of how good the resulting proposed duplicates were.

These results are given in Table \ref{tab:results}
\begin{table}[h]
 \caption{Size of Matching Groups}
  \begin{tabular}{llll}
    \toprule
    Type                        & Epsilon & Max Group Size & No. Match Groups \\
    \midrule
    Traditional Entity Matching &         & 739            & 4,122  \\
    Cosine similarity           & 0.1     & 206            & 2,422  \\
    Cosine similarity           & 0.05    & 72             & 1,161  \\
    L2 norm                     & 0.5     & 240            & 2,959  \\
    L2 norm                     & 0.25    & 51             & 735    \\
    L2 norm                     & 0.125   & 51             & 156    \\
    Cosine similarity           & 0.025   & 51             & 589    \\
    \bottomrule
  \end{tabular}
  \label{tab:results}
\end{table}

\subsection{Musicbrainz Experiments}
A publicly available dataset that was generated for testing Entity Matching was used. These datasets are provided by the University of Leipzig and can be seen at their website\cite{emDatasets}. We used the  \href{https://dbs.uni-leipzig.de/files/datasets/saeedi/musicbrainz-200-A01.csv.dapo}{musicbrainz-200K} dataset for our experiments. This dataset contains 193,750 tuples, where each tuple represents a certain audio recording. The size of the generated duplicate clusters ranges from 1 to 5.

\subsubsection{Match Sentence}
The match sentences were created by concatenating the following fields with spaces:

\begin{itemize}
\item title
\item length
\item artist
\item album
\item year
\item language
\end{itemize}

An example record is shown in Table \ref{tab:examplerecord}

\begin{table}[h]
 \caption{Example Record}
  \begin{tabular}{|l|l|l|l|l|l|}
    \hline
    \textbf{title} & \textbf{length} & \textbf{artist} & \textbf{album} & \textbf{year} & \textbf{language} \\
    \hline\hline
    009-Ballade a donner & 4m 2sec & Luce Dufault & Luce Dufault (1996) & 96 & French \\
    \hline
  \end{tabular}
  \label{tab:examplerecord}
\end{table}

With this example record the "Match Sentence" becomes:

\colorbox{shadecolor}{009-Ballade a donner 4m 2sec Luce Dufault Luce Dufault (1996) 96 French}

\subsubsection{Success Metric}
In order to be able to evaluate the success or otherwise of the generated match groups we used the following metrics:

\begin{itemize}
\item True positives (TP): Correctly declared duplicates
\item False positives (FP): Incorrectly declared duplicates
\item True negatives (TN): Correctly avoided pairs
\item False negatives (FN): Missed duplicates
\item precision = TP / (TP + FP)
\item recall = TP / (TP + FN)
\item F-score = 2 * (precision * recall) / (precision + recall)
\end{itemize}

An initial baseline was run against the dataset using the traditional NLP approach. The records were first sorted on the 'artist' attribute which was then used for blocking. The similarity metric used was the Levenshtein Distance algorithm on the 'title' attribute, with a similarity threshold value of 0.9. This method is referred to as the 'NBA1' when showing the results later.

Various experiments were then run using the proposed technique and different values of epsilon.

\subsubsection{Results}
It was found that actually the cosine similarity distance metric produced good match groups. The l2 distance metric did not yield any good results with the Musicbrainz dataset. Using the cosine similarity distance metric with varying epsilon values produced the following best matching results.

\begin{table}[h]
 \caption{Clustering Results}
  \begin{tabular}{|l|l|l|l|l|l|}
    \hline
    \textbf{Epsilon} & \textbf{Dup} & \textbf{TP} & \textbf{FP} & \textbf{FN} & \textbf{F-score} \\
    \hline\hline
    0.175 & 143,750 & 43,759 & 54,318 & 45,673 & 0.47 \\
    \hline
    0.245 & 143,750 & 59,272 & 42,261 & 42,261 & 0.58 \\
    \hline
    0.265 & 143,750 & 59,029 & 37,361 & 47,360 & 0.58 \\
    \hline
    0.3250 & 143,750 & 42,890 & 25,422 & 75,438 & 0.46 \\
    \hline
  \end{tabular}
  \label{tab:proposedmethodtable}
\end{table}

\begin{figure}[h]
\includegraphics[width=\textwidth]{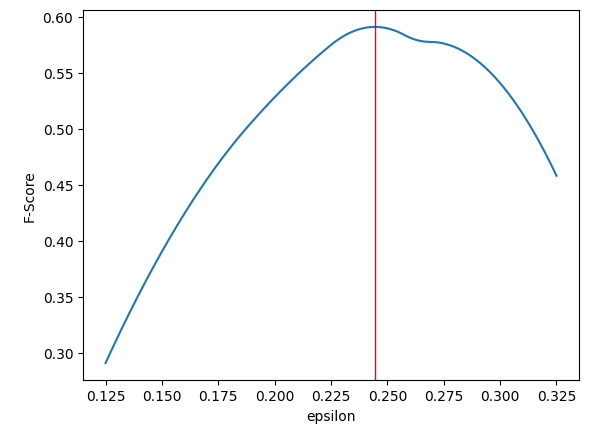}
\caption{Experimental results epsilon against F-score}
\label{fig:epsilonagainstfscore}
\end{figure}

When a graph of the epsilon values against the F-scores was produced, it was possible to see a curve with an obvious maximum value. See Figure \ref{fig:epsilonagainstfscore}

Comparing the results from the traditional NLP method (with the name NBA1 in the table) with the proposed method we can see a massive improvement in the accuracy of the clustering groups that are produced. There is an interesting article that was produced in June 2023 \cite{improvingClustering} that talks about different techniques that can be used to improve the clustering capabilities of the traditional NLP method. These improved results have also been included in this table to show that even using some or all of these techniques, our proposed method is still producing comparably good clusters.

See Table \ref{tab:comparisontable}.
\begin{table}[h]
 \caption{Clustering Methods Results}
  \begin{tabular}{|l|l|l|l|l|l|}
    \hline
    \textbf{Clustering Method} & \textbf{Dup} & \textbf{TP} & \textbf{FP} & \textbf{FN} & \textbf{F-score} \\
    \hline\hline
    NBA1 & 1,201,486 & 29,357 & 1,172,129 & 114,393 & 0.04 \\
    \hline
    SNM1 & 59,757 & 30,737 & 29,020 & 113,013 & 0.3 \\
    \hline
    NBA2 & 66,052 & 53,029 & 13,023 & 90,721 & 0.5 \\
    \hline
    SNM2 & 79,134 & 60,438 & 18,696 & 83,312 & 0.54 \\
    \hline
    \textbf{ProposedMethod} & 143,750 & 59,272 & 42,261 & 42,261 & \textbf{0.58} \\
    \hline
  \end{tabular}
  \label{tab:comparisontable}
\end{table}

The description of the various Clustering Methods are:
\textbf{NBA1}: Naive Blocking Algorithm;
\textbf{SNM1}: Sorted Neighbourhood Method;
\textbf{NBA2}: Naive Blocking Algorithm with pre-processing;
\textbf{SNM2}: Sorted Neighbourhood Method with pre-processing.

\begin{figure}[h]
\includegraphics[width=\textwidth]{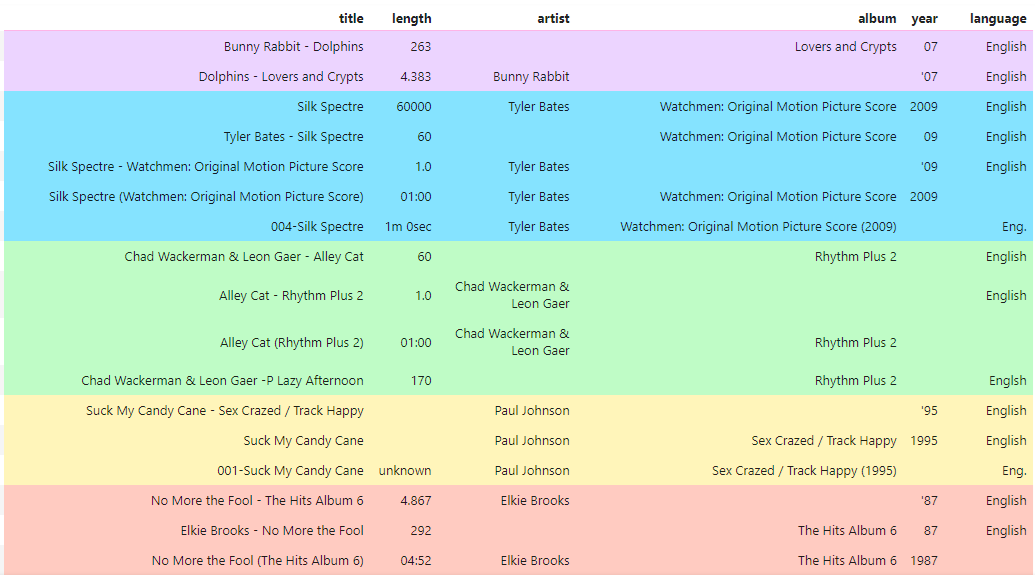}
\caption{Match Groups for the Musicbrainz 200K dataset (epsilon=0.245)}
\label{fig:matchgroups}
\end{figure}

Example cluster groups are shown in Figure \ref{fig:matchgroups}

\subsection{Visualising Clustering}
\begin{figure}[h]
\includegraphics[width=\textwidth]{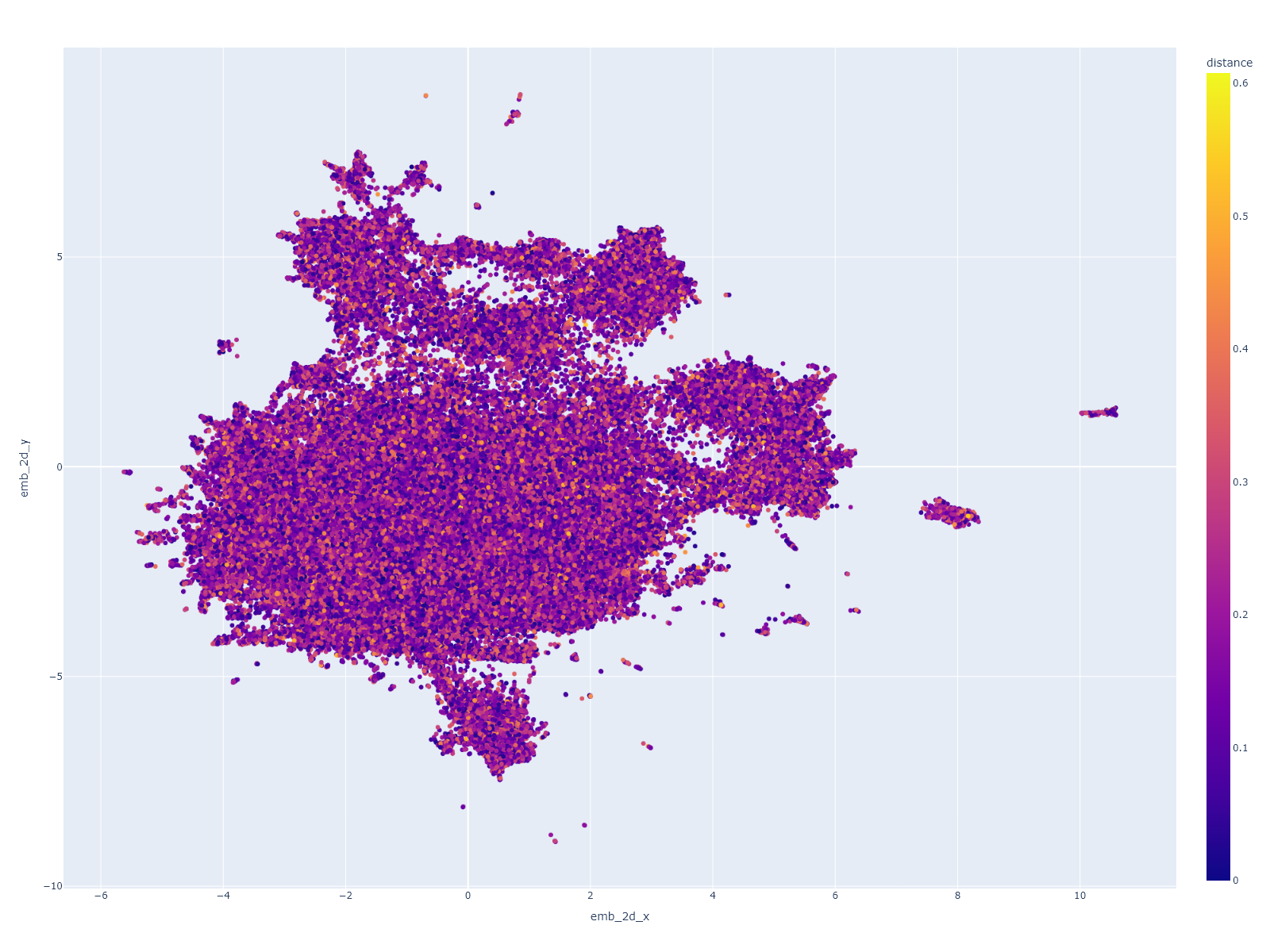}
\caption{2D UMAP Musicbrainz 200K nearest neighbour plot}
\label{fig:embedding2d}
\end{figure}

A useful way to visual the way the records are clustered is to first calculate for each record the distance to it’s nearest neighbour. This was done using the cosine similarity distance metric. Once this was completed the embedding vectors of 768 dimensions were then reduced to just 2 dimensions by using a UMAP \cite{2018arXivUMAP} algorithm provided by the \textbf{umap-learn}\cite{umap-learn} PyPi package. These then became [X, Y] co-ordinates on a scatter plot with each point coloured using the distance metric. The darker the colouring the closer the point is to another. See Figure \ref{fig:embedding2d}

\section{Benefits of the proposed approach}
There are a number of benefits that the proposed approach has over the traditional approach. These are the following:

\subsection{No training of a model required}
Because we are using a pretrained Large Language Model to produce our vectors, there is no need to do further training as we proceed. We already start with good potential match groups.

\subsection{No pre-processing of the data required}
In order to improve the identification of potential matches, pre-processing of the data is often performed. These include filling in missing values, sorted with nearest neighbours and other such techniques. All of these techniques can improve the matching, but also introduce errors. With the proposed approach no pre-processing of the data is required, which in turn means that no errors are introduced.

\subsection{User Engagement}
Because our initial proposed match groups seem already reasonable to the users, this engages them from the beginning. Using the traditional approach means that the initial proposed match groups aren't very good and until the users label them we aren't then able to train a model and propose better candidates. By this point user engagement and confidence is low.

\subsection{Adaptable}
As bigger and better LLMs are produced, these can easily be swapped with the current LLM being used. There will be no need to change the underlying code.

\subsection{Understands Language}
Because of the way that the LLMs have been trained, they understand language. Things such as abbreviations and substitutions are easily picked up, without any pre-processing of the data required. Things such as "and" and "\&" in English are known to be equivalent. The same in other languages such as German where "und", "U." and "\&" are known to be equivalent.

\section{Future Work}
As more and more embedding models are brought out which work over an increasingly larger vector space, it would be good to see how this improves our ability to cluster similar records together. The assumption would seem to be that a larger vector space would ensure a greater ability to cluster similar records together. But it would be good to check that this did indeed turn out to be the case.

It will be beneficial to research an approach to calculate the optimal epsilon based on the distribution of the matching records.

\section{Conclusion}
We have introduced an innovative approach to identifying duplicate records that is comparatively better than the traditional NLP approach, even with various improvements and pre-processing of data that can be done. The proposed approach has the advantage of not requiring any pre-processing of the data to be done before it can be used. Pre-processing of the data can improve accuracy but also introduces its own errors. Another advantage of the proposed approach is that it does not require labelled data to work better, or the training of a model. It also understands differences in language. As bigger and better LLMs are created these can easily be swapped for the current LLM without any changes to underlying technique.

\section{Resources}
Useful documentation and Jupyter notebooks have been made available that explain this approach in detail using the Musicbrainz test data. This is available at \cite{genai-dedup-resources}.

\section*{Acknowledgments}
I would like to thank Dr. Ben Hoyle, a fellow worker at Zeiss Digital Partners, who has helped me with this paper in many different ways. He has been involved in the production of over 50 papers and his experience and guidance has been invaluable. I really couldn't have done this without his input. Thank you Ben.

\bibliographystyle{unsrt}  
\bibliography{references}  

\section*{Biography Of Author}
\begin{wrapfigure}{L}{21mm}
\vspace{-\intextsep}
\includegraphics[width=21mm]{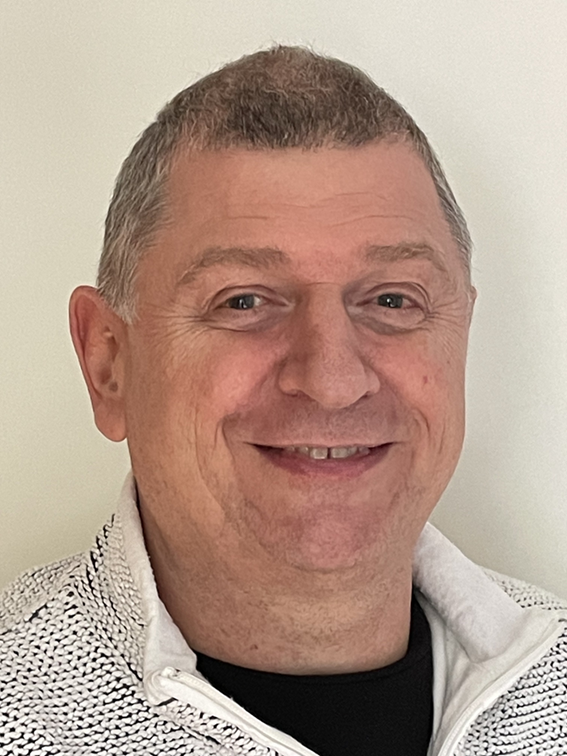}
\end{wrapfigure} 
\textbf{Ian Ormesher} \orcidlink{0009-0008-4041-9443} is a seasoned full-stack Data Scientist with a robust background in training and deploying AI models in production environments. With a career spanning over four decades in the computer industry, he has honed his skills in Machine Learning, Deep Neural Networks, NLP, GenAI, Reinforcement Learning, and Computer Vision. He is proficient in a wide array of programming languages and data analysis tools.

\end{document}